%% file: root.tex
\documentclass[a4paper,conference]{IEEEtran}

\usepackage{cite}
\usepackage{graphics}
\usepackage{epsfig}
\usepackage{amsmath} 
\usepackage{amssymb}
\usepackage{multirow}
\usepackage{makecell}
\usepackage{adjustbox}
\usepackage{subcaption}
\usepackage{float}
\usepackage{soul}
\usepackage{xcolor}
\usepackage{diagbox} 
\usepackage{algorithm}
\usepackage{mathtools}
\usepackage{algorithmicx}
\usepackage{algpseudocode}
\usepackage{listings}
\usepackage{ctable}
\usepackage[nolist,nohyperlinks]{acronym}
\usepackage{hyperref}
\usepackage{pifont}

\acrodef{ml}[ML]{Machine Learning}
\acrodef{dl}[DL]{Deep Learning}
\acrodef{cnn}[CNN]{convolutional neural network}
\acrodef{sae}[SAE]{supervised autoencoder}
\acrodef{sd}[SD]{Small Data}
\acrodef{sasd}[SASD]{supervised autoencoder with self-supervised decoders}
\acrodef{gan}[GAN]{generative adversarial network}
\acrodef{ntk}[NTK]{neural tangent kernel}
\acrodef{cntk}[CNTK]{convolutional neural tangent kernel}

\begin{document}

\title{A Close Look at Deep Learning with Small Data}

\author{\IEEEauthorblockN{Lorenzo Brigato}
\IEEEauthorblockA{Dpt. of Computer, Control, and Management\\
Engineering\\
Sapienza University of Rome\\
Rome, Italy\\
brigato@diag.uniroma1.it}
\and
\IEEEauthorblockN{Luca Iocchi}
\IEEEauthorblockA{Dpt. of Computer, Control, and Management\\
Engineering\\
Sapienza University of Rome\\
Rome, Italy\\
iocchi@diag.uniroma1.it}
}

\maketitle

\begin{abstract}
In this work, we perform a wide variety of experiments with different deep learning architectures on datasets of limited size.
According to our study, we show that model complexity is a critical factor when only a few samples per class are available.
Differently from the literature, we show that in some configurations, the state of the art can be improved using low complexity models.
For instance, in problems with scarce training samples and without data augmentation, low-complexity convolutional neural networks perform comparably well or better than state-of-the-art architectures.
Moreover, we show that even standard data augmentation can boost recognition performance by large margins.
This result suggests the development of more complex data generation/augmentation pipelines for cases when data is limited.
Finally, we show that dropout, a widely used regularization technique, maintains its role as a good regularizer even when data is scarce.
Our findings are empirically validated on the sub-sampled versions of popular CIFAR-10, Fashion-MNIST and, SVHN benchmarks.

\end{abstract}


\IEEEpeerreviewmaketitle

\input{introduction}
\input{relatedwork}

\input{problem}

\input{results}

\input{conclusion}

\bibliographystyle{IEEEtran}
\bibliography{biblio}

\end{document}

%% file: introduction.tex
\section{Introduction}
\label{sec:introduction}


\ac{ml} popularity has rapidly increased thanks to the success of deep learning \cite{DBLP:journals/nature/LeCunBH15}.
In particular, \ac{cnn} architectures, have achieved considerable success in a wide range of computer vision tasks
including object classification \cite{he2015delving}, object detection \cite{redmon2016you} and semantic segmentation \cite{long2015fully}, just to cite a few.
The two main ingredients that have favored the rise of this type of algorithms are i) networks with deeper structure and ii) the use of large annotated datasets.
The latter requirement can not always be fulfilled for several reasons.
Obtaining and labeling data is needed to achieve strong results but this process might be extremely expensive or not possible at all.
For instance, in the medical field, high-quality annotations by radiology experts are often costly and not manageable at large scales \cite{litjens2017survey}.
Several sub-domains of \ac{ml} are trying to mitigate the necessity of training data, tackling the problem from different perspectives.
Transfer learning aims at learning representations from one domain and \textit{transfer} the learned features to a closely related target domain \cite{pan2009survey}, \cite{bengio2012deep}.
Similarly, few-shot learning uses a \textit{base set} of labelled pairs to generalize from a scarce \textit{support set} of target classes \cite{vanschoren2018meta}.
Both approaches have gained much attention in the community but still require a large source of annotated data.
Furthermore, the target domain must be related to the source domain.
Another research direction is trying to reduce the demand for annotations.
This field is known as self-supervised learning.
Usually, a large pool of images is used to teach how to solve a \emph{pretext} task to a \ac{cnn} \cite{jing2020self}.
This task does not require human annotations and is conceived to teach general visual features that can be transferred to the \textit{downstream} task.
In this manner, costly human annotations are not needed but there is still the problem of collecting many images.
In general, we would like to have systems that can recognize objects from just a few exemplars.

In this work, we present a detailed empirical study of deep learning models in the small data regime.
Similarly to what has been done in \cite{barz2020deep} and \cite{DBLP:conf/iclr/AroraD0SWY20}, we benchmark the approaches by varying the number of data points in the training sample while keeping it low with respect to the current standards of computer vision datasets.
Such a scenario can also be seen as the starting periods of an application that collects data over time and aims to perform the task in the best possible way before it has collected large amounts of data.
For example, a robot involved in an interactive activity with people needs to recognize human actions or behaviour \cite{zhang2019prediction}, \cite{valipour2017incremental}.
Due to its inherent difficulty, not many past works have studied such a problem.
Indeed, it is very hard to train function approximators when the availability of multi-dimensional points to interpolate is scarce.
Any \ac{ml} model is highly prone to overfit the training dataset, especially if its complexity is too high to handle the current task.
\ac{cnn}s have proved to be extraordinarily resistant to overfitting, although the number of trainable parameters is usually much greater than the number of data points \cite{kawaguchi2017generalization}.
In this paper, we show that model complexity is a critical factor in small-data domains and that small nets can be better than big ones in scenarios with limited training samples.
A large number of works have proposed techniques to increase the generalization capabilities of deep networks.
For instance, data augmentation is commonly used to fight the data scarcity problem and to increase generalization \cite{shorten2019survey}.
We perform our analysis with and without data augmentation to understand its effectiveness when the dataset size is limited.
In our experiments, we show that standard augmentation pipelines improve recognition performance up to large margins.
Clearly, the augmentation type should be carefully designed since its success is highly correlated to the image type.
Moreover, dropout is an extremely popular technique that regularizes neural networks \cite{srivastava2014dropout} by randomly dropping units to prevent co-adaptation and to favor generalization through ensembles.
In this paper, we show that dropout is a good regularizer even when data is scarce, slightly less effective when the number of samples per class is extremely low (e.g. $10$).
In summary, the contributions of this paper are the following: \emph{1)} we perform a large set of experiments with different quantities of training data and deep architectures over three popular computer vision benchmarks; \emph{2)} we show that for small-data problems, low-complexity \ac{cnn}s are comparable to or better than high-complexity ones depending on the training set dimension and the use of data augmentation; \emph{3)} we demonstrate that standard data augmentation consistently improves testing accuracy of deep networks trained with few samples; \emph{4)} we show that dropout results to be a good regularizer even in these small-data settings.

\begin{table*}
	\centering
	\input{tables/parameters}
	\caption{In this table, we give more details regarding the computational complexity of the tested networks. We show the number of convolutional filters per layer of the standard \ac{cnn}s used. Further, we report for each model and dataset the number of trainable parameters (PARAMS) and floating-point operations (FLOPs), two standard metrics to measure model complexity.}
	\label{tab:models_structure}
\end{table*}

%% file: tables/parameters.tex
\begin{tabular}{lllllllllll}
		\toprule
   & \multicolumn{4}{c}{Filters} & \multicolumn{2}{c}{CIFAR-10} & \multicolumn{2}{c}{FMNIST} & \multicolumn{2}{c}{SVHN}\\

\cmidrule(lr){2-5} \cmidrule(lr){6-7} 	\cmidrule(lr){8-9}  	\cmidrule(lr){10-11}  
\textbf{Model} & c1 & c2 & c3 & c4 & PARAMS     & FLOPs    & PARAMS     & FLOPs      & PARAMS     & FLOPs\\

\midrule

CNN-lc     & 8 & 16 & 32 & 64    & $2.7 \cdot 10^{4}$     & $5.4 \cdot 10^{4}$     & $2.5 \cdot 10^{4}$    & $4.9 \cdot 10^{4}$     & $2.7 \cdot 10^{4}$      & $5.4 \cdot 10^{4}$     \\

CNN-mc     & 16 & 32 & 64 & 128    & $1.02 \cdot 10^{5}$     & $2.05 \cdot 10^{5}$     & $9.8 \cdot 10^{4}$    & $1.96 \cdot 10^{5}$  & $1.02 \cdot 10^{5}$      & $2.05 \cdot 10^{5}$    \\

CNN-hc      & 32 & 64 & 128 & 256   & $3.98 \cdot 10^{5}$     & $7.96 \cdot 10^{5}$     & $3.9 \cdot 10^{5}$    & $7.79 \cdot 10^{5}$      & $3.98 \cdot 10^{5}$     & $7.96 \cdot 10^{5}$     \\

ResNet-20  &  &  &  &     & $2.73 \cdot 10^{5}$     & $1.62 \cdot 10^{6}$     & $2.72 \cdot 10^{5}$    & $1.62 \cdot 10^{6}$     & $2.73 \cdot 10^{5}$      & $1.62 \cdot 10^{6}$     \\
\bottomrule  
                                        
\end{tabular}

%% file: relatedwork.tex
\section{Related Work}
\label{sec:related_work}

Learning from datasets of limited size is extremely challenging and, for this reason, largely unsolved.
As previously said, few works have tried to tackle the problem of training deep architectures with a small number of samples due to the difficulty of generalizing to novel instances.\\
We start by mentioning a series of works that focused on the classification of vector data and mainly used the UCI Machine Learning Repository as a benchmark.
In \cite{fernandez2014we} the authors have shown the superiority of random forests over a large set of classifiers including feed-forward networks.
Later, \cite{olson2018modern} used a linear program to empirically decompose fitted neural networks into ensembles of low-bias sub-networks. 
They showed that these sub-networks were relatively uncorrelated which lead to an internal regularization process similar to what happens in random forests, obtaining comparable results.
More recently, \cite{DBLP:conf/iclr/AroraD0SWY20} proposed the use of \ac{ntk} architectures in low data tasks and obtained significant improvements over all previously mentioned classifiers.\\
All previous works did not test \ac{cnn}s since inputs were not images.
When the input dimensionality increases, the classification task inevitably becomes more complex.
For this reason, a straightforward approach to improve generalization is to implement techniques that try to synthesize new images through different transformations (e.g. data augmentation \cite{shorten2019survey}).
Some previous knowledge regarding the problem at hand might turn to be useful in some cases \cite{hu2017frankenstein}. However, this makes data augmentation techniques not always generalizable to all possible image classification domains.
It has also been proposed to train generative models (e.g. GANs) to increase the dataset size and consequently, performance \cite{liu2019generative}.
Generating new images to improve performance is extremely attractive and effective. Yet, training a generative model might be computationally intensive or present severe challenges in the small sample domain.
In our work, we use standard data augmentation and do not focus on approaches that improve image synthesis.
We show that even standard augmentation can be highly effective to improve recognition performance and confirm that data augmentation/generation is a promising approach for datasets of limited size.\\
\cite{DBLP:conf/iccci/Rueda-PlataRG15} suggested to train \ac{cnn}s with a greedy layer-wise method, analogous to that used in unsupervised deep networks and showed that their method could learn more interpretable and cleaner visual features.
\cite{barz2020deep} proposed the use of the cosine loss to prevent overfitting claiming that the L2 normalization involved in the cosine loss is a strong, hyper-parameter-free regularizer when the availability of samples is scarce.
They obtained the best results on fine-grained datasets that have between $20$ and $80$ samples per class.
On the other hand, \cite{DBLP:conf/iclr/AroraD0SWY20} performed experiments with \ac{cntk} networks on small CIFAR-10 showing the superiority of \ac{cntk} in comparison to a ResNet-34 model.
Finally, \cite{bornscheinsmall} studied the generalization performance of deep networks as the size of the training set varies.
They found out that even larger networks can handle overfitting and obtain comparable or better results than smaller nets if properly optimized and calibrated.

%% file: problem.tex
\section{Small-Data Classification Problem}
\label{sec:models}

In this section, we first outline the definition of our classification problem with limited data, followed by describing the deep learning architectures studied in our experiments. Finally, we describe the datasets and regularization techniques used in this work.

\subsection{Problem definition}
As a standard supervised \ac{ml} problem, given a training distribution of images $\mathcal{X}$ and a label distribution $\mathcal{Y}$, our objective is to learn a classifier $f_{\theta}$, parametrized by a set of variables $\theta$, such that for any image $x \sim \mathcal{X}$ with corresponding label $y \sim \mathcal{Y}$, $y = f_{\theta}(x)$.
Let us suppose that our distribution is made of $K$ different classes and that our training dataset sampled from $\mathcal{X}$ and $\mathcal{Y}$, is composed of $N$ images per class.

Differently from the standard computer vision datasets, we choose to keep $N$ low.
Despite the notion of low is highly subjective, we follow two different sub-sampling protocols for setting our experiments.
Firstly, to directly compare our tested models with the networks proposed in \cite{DBLP:conf/iclr/AroraD0SWY20} $N$ is doubled each time starting from $N_{min} = 1$ up to $N_{max} = 128$.
Secondly, since we were also interested in configurations with more data, we also varied $N$ in the set $\{10, 20, 40, 80, 160, 320, 640, 1280\}$, corresponding to a ten-time increase of samples per class with respect to the just cited protocol. 
In this manner, it is easy to benchmark models capabilities as the number of samples per class increases.
In the result section, we specify when we are using the first or second sub-sampling protocol.

As previously said, such a protocol could be easily found in a situation where data is collected through time. Understanding which is the more effective model before reaching large quantities of data is an important research question.

In general, there is no specific limit that regards the number of classes $K$.
In our work, we have chosen three problems with $K = 10$.
  
\subsection{Models}
Since we are interested in comparing deep models with different complexity we define three \ac{cnn}s with increasing filter widths and a more complex ResNet-20.\\\
\textbf{\ac{cnn}.} We test a standard \ac{cnn} architecture made of convolutional and max-pooling layers as feature extractors. The entire network minimizes the standard classification loss:
\begin{equation}
\label{eq:cl_loss}
    \frac{1}{t} \sum_{i=1}^{t} L_{c}(y_{i}, f_{\theta}(x_{i}))
\end{equation}
More details about the structure will follow.\\
\textbf{ResNet-20.} Residual networks were introduced in \cite{he2016deep} to improve the training of very deep neural networks.
The basic learning block of such networks is the residual block.
ResNets are made of different blocks of stacked layers (convolutions with non-linear activation functions and batch normalization). Shortcut connections join the input of each block to the output. 
This addition helps gradients to flow backward and ease the training of the overall network.
ResNet-20 minimizes the same loss described in Eq. \ref{eq:cl_loss}.\\
\textbf{Models structure.} The \ac{cnn}s process the input image through four convolutional layers. 
The kernel is of size $3$ and moves with stride $1$ on both width and height directions.
On the other hand, the max-pooling layers have a pool-size and stride of $2$.
The output of the convolutional feature extractor is flattened.
Then, a feed-forward layer maps the extracted features to the requested output dimensionality $K$.\\
We do not describe the full structure of ResNet-20 that can be found in \cite{he2016deep}. For what concerns the kernel widths of ResNet-20, we set them to $16$, $32$ and $64$ as in the default CIFAR-10 network proposed in \cite{he2016deep}.\\
\textbf{Models complexity.} We control the complexity of the \ac{cnn}s by the number of filters in convolutional layers.
We fix the lowest model complexity for the \ac{cnn} by setting a number of filters for the first convolutional layer.
Then, for each deeper layer, we simply double their number in order to increase complexity.
We define a low, medium and high complexity \ac{cnn} that we will call \ac{cnn}-lc, \ac{cnn}-mc, \ac{cnn}-hc.
The three architectures have respectively $8$, $16$ and $32$ base filters.
A detailed description can be found in Tab. \ref{tab:models_structure}.
We also report the complexity in terms of parameters and floating-point operations (FLOPs) of all networks. \\
\textbf{Training setup.} All standard \ac{cnn}s have been optimized with Adam and default parameters \cite{DBLP:journals/corr/KingmaB14}.\\
For what concerns ResNet-20, we slightly modified the training policy originally proposed in \cite{he2016deep}.
The optimizer used is stochastic gradient descent (SGD) with weight decay and Nesterov momentum respectively set to \(10^{-4}\) and \(0.9\).
We start with a learning rate of \(0.1\) and decrease it after 75\% of the total number of iterations by an order of magnitude.
We increased the number of iterations with the initial learning rate to be sure of decreasing it after having reached the training loss plateau. Furthermore, we have noticed that decreasing a second time the learning rate did not improve further the testing performance. 
The network was fed with a mini-batch of \(32\) images.
We preferred a smaller batch with respect to the original \(128\) since led to better performance.
We also trained the other \ac{cnn}s with a batch size of $32$ for the same reason.
Due to different training set dimensions (including the use of data augmentation), we train the networks for a different number of epochs.
For the sub-sampling protocol with $N_{min} = 1$, assuming that the first element of the set corresponds to datasets with $N_{min}$ samples per class and the last one to $N_{max}$, architectures are trained for $\{600,  600, 600, 400, 325, 275,  225, 175\}$.\\
For the other sub-sampling protocol ($N_{min} = 10$) and without data augmentation, models are trained for $\{400, 300, 250, 200, 150, 100, 50, 50\}$ epochs.
When augmentation is used, we slightly increase the number of epochs to be sure that models have converged ($\{400, 350, 300, 250, 200, 150, 100, 100\}$).
Note that standard \ac{cnn}s require far fewer iterations than ResNet-20 to converge because of their inherent lower complexity and the optimizer used.
Indeed, Adam is known to be faster than SGD in terms of converge speed.
To be sure that overfitting did not influence our results, for each run, the maximum value of the accuracy scored on the test set was considered.
We mainly use the categorical cross-entropy as classification loss ($L_{c}$).
When specified, we perform some comparisons with the cosine loss proposed in \cite{barz2020deep}.

\subsection{Datasets}

We perform our experiments on three popular computer vision datasets, namely CIFAR-10 \cite{krizhevsky2009learning}, FMNIST (Fashion-MNIST) \cite{xiao2017fashion} and SVHN \cite{netzer2011reading}.\\
FMNIST is a popular dataset comprised of black and white images of dimension \(28 \times 28\).
The total number of categories is \(10\) and each class belongs to fashion items including dresses, trousers, sandals, etc. 
The dataset has originally \(50,000\) training images and \(10,000\) testing images with both sets balanced.\\
CIFAR-10 is an established computer vision benchmark consisting of color images (\(32 \times 32\)) coming from \(10\) different classes of objects and animals.
The sizes of training and test splits of CIFAR-10 are equal to the ones of FMNIST.\\
Finally, SVHN is a real-world image dataset semantically similar to MNIST since contains images of digits.
However, it is significantly harder because recognizing numbers in natural scene images constitutes a more complex task.
The cropped-version sets used originally have $73,257$ training and $26,032$ testing images of dimension $32 \times 32 \times 3$.\\
We have chosen datasets made of ten classes and relatively small images since the restrictions imposed on the number of samples per class make the problem already challenging.
As previously anticipated, since all three original datasets contain several samples per class (many more than our defined $N_{max}$), we build, for each dimension, a sub-sampled version of the original dataset.
We will refer to any down-sampled version of any dataset as $sDatasetName-n\_samples$.
For instance, sCIFAR-10-20 indicates the down-sampled version of CIFAR-10 with twenty samples per class.\\
The testing sets remain fixed to the original sizes throughout all the evaluations.
To ensure consistency of the results, we perform \(10\) runs for each sub-sampled version of each dataset.

\subsection{Regularization techniques}
\label{subsec:regularizations}

We study the influence of two popular regularization techniques for some of our experiments with datasets of limited size.\\
\textbf{Dropout.} We use dropout in the last layer of \ac{cnn}-lc, \ac{cnn}-mc and \ac{cnn}-hc.
We vary the dropping-rate probabilities in the set $\{0.0, 0.4, 0.7\}$, corresponding to absent, medium and, high dropout.
We will show that high rates help to prevent overfitting except for rare cases.\\
\textbf{Data augmentation.} Artificially increasing the images of a dataset by applying data augmentation techniques has been shown to improve the generalization capabilities of deep models even when a great quantity of data is available \cite{shorten2019survey}.\\
The transformations applied to the images should not modify the semantics needed to perform the classification task.
To this end, we applied three different sets of augmentations to the three datasets.
Inputs are augmented at each epoch, constituting a \textit{non-static} dataset.\\
For CIFAR-10 images we used the standard augmentation pipeline also used in \cite{he2016deep}.
More precisely, each image is padded 4 pixels on each side and a \(32 \times 32\) crop is randomly sampled from the padded image or its horizontal flip.\\
FMNIST objects present small variations in orientation and scale with respect to the ones of CIFAR-10.
In this case, we padded the images with 2 pixels per side and cropped a window of \(28 \times 28\) pixels.
We have not applied horizontal flipping.\\
Finally, digits in SVHN undergo noticeable changes in terms of illumination and contrast.
For this reason, we adjusted image brightness and contrast by random factors.
The first one was contained in $\pm 0.2$ while the second one in between $0.2$ and $2.0$.
Further, we applied the cropping policy of CIFAR-10 without considering the horizontal flipping that could change the numbers semantics.

%% file: results.tex
\section{Results}

\begin{table*}[t]
	\centering
	{\addtolength{\tabcolsep}{-4pt}{\scriptsize\input{tables/all_results}}}
	\caption{Average testing accuracy for all tested models considering the sub-sampling protocol with $N_{min} = 10$ and $N_{max} = 1280$.
	The best value per column is reported in bold.
	Note that only in some cases we used data augmentation.
	The numbers below the \textit{Dropout} column stand for the probability of dropping units for the dropout layer.
	We do not report standard deviations to improve table readability.
	If needed, standard deviations are qualitatively shown in the figures that analyze the proposed comparisons of our paper.}
	
	\label{tab:all_results}
\end{table*}

\subsection{Influence of models complexity on performance}
\label{subsection:model_complexity}

\begin{figure*}[t]
	\centering
	\includegraphics[scale=0.4]{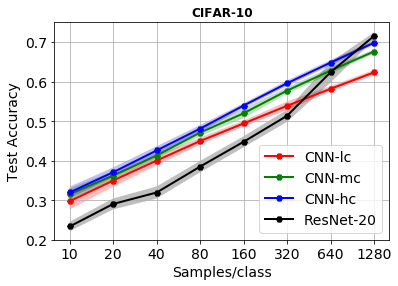}
	\centering	\includegraphics[scale=0.4]{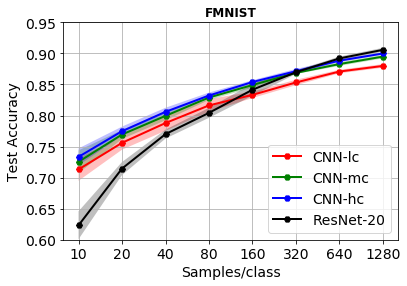}
	\centering	\includegraphics[scale=0.4]{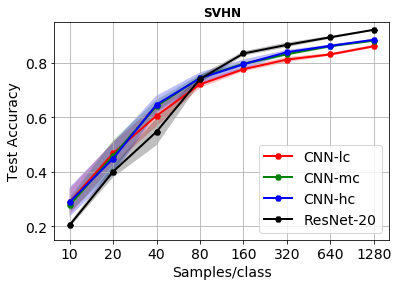}
	\caption{Comparison between networks with different complexities. The simpler \ac{cnn}s are all equipped with dropout (dropping rate equal to $0.7$). In these experiments, data augmentation is not used. Results, over the test sets, are in terms of accuracy, averaged over $10$ runs.}
	\label{fig:model_complexity}
\end{figure*}

First, we analyze the impact of model complexity in our small-data classification problems.
For this reason, we have chosen to test the three standard \ac{cnn}s with increasing complexity (\ac{cnn}-lc, \ac{cnn}-mc, and \ac{cnn}-hc) and ResNet-20 that has the highest complexity in terms of FLOPs (check Tab. \ref{tab:models_structure} for more details).
In this case, all networks are trained with the cross-entropy loss.
The results of this analysis are shown in Fig. \ref{fig:model_complexity}.
Note that for this analysis, data augmentation is not used.
The standard \ac{cnn}s consistently outperform ResNet-20 in the sCIFAR10 problem when $N$ is smaller than $320$ samples.
ResNet scores the best accuracy out of the four models only when $N = 1280$.
The difference between ResNet-20 and \ac{cnn}-hc is roughly $10\%$ for sCIFAR10-10 up to sCIFAR10-320.
For larger training sets, the difference sharply decreases until becomes negative.
It is interesting to note that also the \ac{cnn}-lc keeps a consistent gap up to sCIFAR10-320 despite it has roughly two orders of magnitudes fewer FLOPs than ResNet-20.
Similar behavior can be noted in sFMNIST and sSVHN with the only difference that the most complex model increases its performance more rapidly.
This is probably due to the intrinsic difficulty of the classification problem.
When the training and testing distributions are more similar (e.g. sFMNIST or sSVHN), the addition of training samples helps the classifier at a faster rate.
In sFMNIST and sSVHN, ResNet starts to have a comparable accuracy when the number of samples per class is, respectively, at least $160$ and $80$.
The gap reaches a net difference in terms of average accuracy of around $10\%$ for sSVHN-40 and sFMNIST-10.
For what concerns the comparison between the \ac{cnn}s, we note that results are comparable although there is a significant difference in terms of model complexity (i.e. \ac{cnn}-hc has more than one order of magnitude further parameters than \ac{cnn}-lc).

\subsection{Influence of regularization techniques on performance}

\begin{figure*}[t]
	\centering
	\includegraphics[scale=0.4]{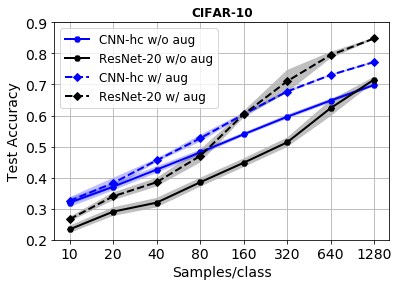}
	\centering	\includegraphics[scale=0.4]{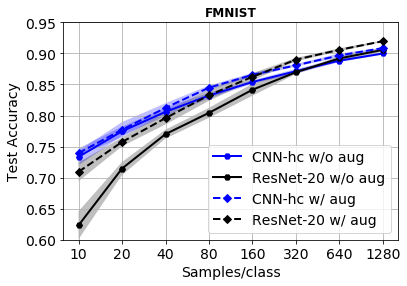}
	\centering	\includegraphics[scale=0.4]{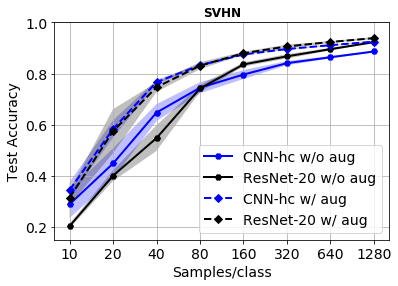}
	\caption{Comparison between \ac{cnn}-hc and ResNet-20 trained with and without data augmentation. \ac{cnn}-hc is using dropout with a dropping rate of $0.7$. Results are in terms of accuracy averaged over $10$ runs.}
	\label{fig:data_aug}
\end{figure*}

\begin{figure*}[t]
  \centering
  \includegraphics[scale=0.35]{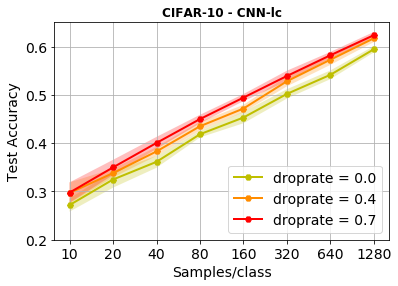}
  \hspace{0.15cm}
  \includegraphics[scale=0.35]{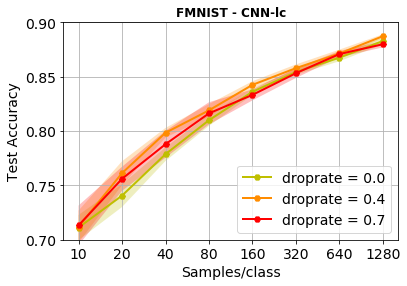}
    \hspace{0.15cm}
  \includegraphics[scale=0.35]{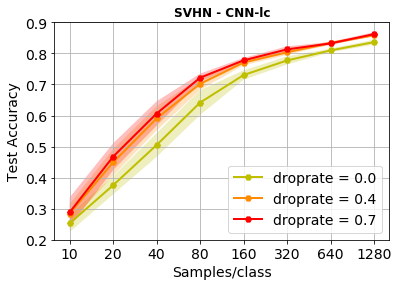}
  \vspace{0.15cm}
  \includegraphics[scale=0.35]{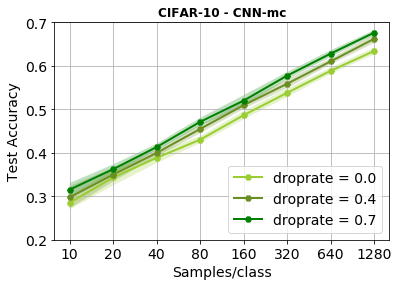}
  \hspace{0.15cm}
  \includegraphics[scale=0.35]{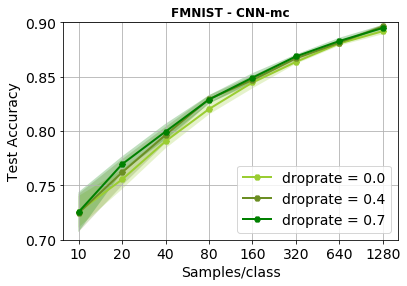}
    \hspace{0.15cm}
  \includegraphics[scale=0.35]{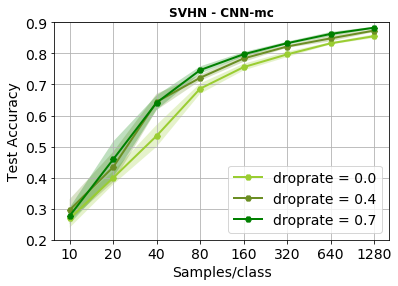}
  \includegraphics[scale=0.35]{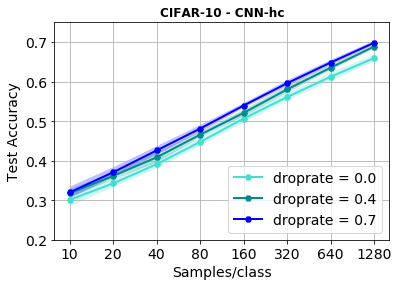}
  \hspace{0.15cm}
  \includegraphics[scale=0.35]{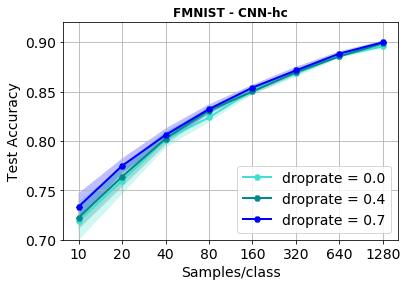}
    \hspace{0.15cm}
  \includegraphics[scale=0.35]{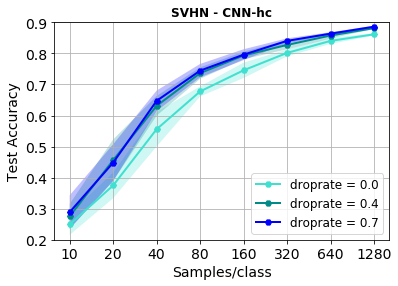}
	\caption{Analysis regarding the influence of dropout on testing accuracy with models of different complexities and three levels of dropout (absent, medium, and high). All networks are trained without data augmentation. Results are in terms of average accuracy obtained over $10$ runs.}
	\label{fig:dropout}
\end{figure*}

Here, we study how two popular regularization techniques such as dropout and data augmentation influence recognition performance when a few samples are available.\\
\textbf{Dropout.} Fig. \ref{fig:dropout} shows the results of the three \ac{cnn}s with different levels of dropout on the three datasets.
Exact numbers are reported in Tab. \ref{tab:all_results}.
For all three cases, dropout regularizes the dense layer of the classifier.\\
In sCIFAR-10, we note that high drop-rates ($0.7$) improve the generalization abilities of all three networks.
Higher improvements over the case without dropout are obtained with the \ac{cnn}-lc (around $5\%$).
However, also \ac{cnn}-hc outperforms its variants with absent and medium dropout.\\
A similar picture can be seen for sSVHN.
Here, improvements are even more noticeable when the training set is limited.
For instance, \ac{cnn}-mc with high dropout outperforms its version without dropout by $\sim 10\%$ in sSVHN-40.
Improvements decrease as the training set increases, but they are still noticeable.\\
Finally, for sFMNIST the improvements are less important, and using or not using dropout does not seem to influence much the final testing accuracy.
When happening, improvements are very small.
The case in which dropout seems to be more effective is when using \ac{cnn}-hc and $10$ or $20$ samples per class.\\
To summarize, dropout maintains its role of good regularizer also when the dataset has a limited size.
Networks equipped with it either generalize better either perform comparably well depending on the type of dataset.\\
\textbf{Data augmentation.} We report the results regarding the use of data augmentation in Fig. \ref{fig:data_aug}. 
Refer to Tab. \ref{tab:all_results} to read the exact values.
We have chosen to train two networks with the previously specified augmentation policies.
The first one is the \ac{cnn}-hc that generally scored the highest testing accuracy among the standard \ac{cnn}s.
The second one is the ResNet-20 which is expected to noticeably improve given the help of data augmentation.
Starting from sCIFAR-10, we note that data augmentation boosts recognition abilities in almost all cases, playing an important role from roughly $160$ samples per class.
The maximum and minimum gains of \ac{cnn}-hc corresponds to $\sim 9\%$ and $\sim 1\%$ with $N = 640$ and $10$.
On the other hand, net improvements of augmented ResNet-20 are $\sim 20\%$ and $\sim 3\%$ with $320$ and $10$ samples per class.
It is interesting to notice that data augmentation allows ResNet-20 to match the \ac{cnn}-hc testing accuracy with $N = 160$.
We have previously seen in the analysis without augmentation that this result was happening given at least $640$ samples per class.\\
Large gains are also obtained on sSVHN.
However, the trend is different from the one seen on sCIFAR-10.
Indeed, in this case, the greatest improvements are obtained with smaller training sets ($ 10 \leq N \leq 160$).
Augmented \ac{cnn}-hc scores a maximum increase of $\sim 14\%$ with $N = 20$ while ResNet-20 of $\sim 20\%$ with $N = 40$.
Differently from sCIFAR-10, we also notice that augmentation does not give a clear advantage to ResNet-20.
Indeed, the two architectures perform comparably well, except for larger datasets ($N > 320$) where ResNet-20 scores improvements of around $1\%$.\\
Finally, data augmentation in sFMNIST shows low gains for \ac{cnn}-hc (maximum around $1.5\%$).
On the other hand, ResNet-20 is still largely helped when data is very scarce.
Testing accuracy is boosted by $\sim 8\%$ in the case of sFMNIST-10.
As we have also seen in sCIFAR-10, here ResNet-20 can match \ac{cnn}-hc performance when trained with smaller datasets.
In this scenario, it happens with half of the samples, more precisely when $N$ is equal to $160$.\\
To sum up, from our experiments we have seen that the success of data augmentation seems to be related to the dataset size and type.
However, in many cases, testing accuracy has been boosted (up to $20\%$ margins) making even standard augmentation policies a must-use when dealing with a small dataset.

\subsection{Comparison with the state of the art}

\begin{figure}[h]
	\centering
	\includegraphics[scale=0.6]{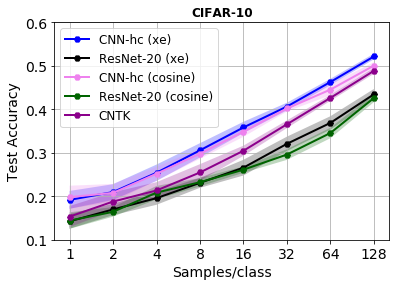}
	\caption{Comparison between \ac{cnn}-hc and ResNet-20 trained with cross-entropy loss and other state-of-the-art techniques. Results are given in terms of accuracy averaged over $10$ runs. Models are trained without any kind of data augmentation.}
	\label{fig:comp_sota}
\end{figure}

Finally, in Fig. \ref{fig:comp_sota} we compare the \ac{cnn}-hc and ResNet-20 trained with the cross-entropy loss to state-of-the-art techniques.
More precisely, we use the sub-sampling protocol of CIFAR-10 originally proposed in \cite{DBLP:conf/iclr/AroraD0SWY20} ($N_{min} = 1$, $N_{max} = 128$) to compare our networks to \ac{cntk}.
In this case, augmentation is turned off to match the procedure used in \cite{DBLP:conf/iclr/AroraD0SWY20}.
Further, we also train \ac{cnn}-hc and ResNet-20 with the cosine loss that was proposed as a state-of-the-art loss function for deep learning with tiny datasets in \cite{barz2020deep}.\\
We start to compare the \ac{cnn}-hc and ResNet-20 trained with the two different losses.
We note that the cosine loss makes the two architectures either comparable to either worst than networks trained with the cross-entropy.
Indeed, the model trained with the cross-entropy loss is more performant in many cases.
We presume that the dataset type, size, and complexity of the model greatly influence the final performance.
The experiments in \cite{barz2020deep} were run with high-capacity neural nets (i.e. ResNet-110, ResNet-50) on mainly fine-grained datasets with a greater number of classes.
Furthermore, in all training set-ups, data augmentation was included.
In this setup proposed by Arora et al. \cite{DBLP:conf/iclr/AroraD0SWY20}, the cosine loss does not outperform the cross-entropy.
This setup is extremely hard for larger networks since the number of training samples is extremely low and data augmentation is not used.
The obtained results are reasonable and somewhat expected.
Indeed, ResNets trained with cross-entropy or cosine loss are the worst models.
More studies regarding the use of the cosine loss should be performed to actually understand when it's advantageous over the cross-entropy.\\
Then, we compare \ac{cnn}-hc and ResNet-20 with \ac{cntk}.
As was already shown in \cite{DBLP:conf/iclr/AroraD0SWY20}, the \ac{cntk} architecture outperforms ResNets.
In the original paper, the comparison was with a ResNet-34.
We confirm the superiority of \ac{cntk} also over a smaller ResNet.
However, we note that our \ac{cnn}-hc clearly outperforms \ac{cntk} in all cases.
The gains are around $\sim 3/5\%$ throughout all training splits with the maximum obtained with $N = 16$ and minimum with $N = 2$.\\
To recap, when using the hard training protocol proposed in \cite{DBLP:conf/iclr/AroraD0SWY20} with extremely limited samples and without data augmentation, the cosine loss does not give a clear advantage over the cross-entropy considering the two main architectures that we have used in this work.
Moreover, the computationally simple \ac{cnn}-hc outperforms \ac{cntk} in all sub-sampled versions of CIFAR-10 making low-complexity \ac{cnn}s
a viable choice in these extreme training conditions.

%% file: tables/all_results.tex
\begin{tabular}{lcccccccccc}
\toprule
                  & \multicolumn{9}{c}{\textbf{Dataset}}\\
\midrule
\textbf{Model}   & \multicolumn{1}{l}{Augmentation}  & \multicolumn{1}{c}{Dropout} & \multicolumn{1}{l}{sCIFAR10-10} & \multicolumn{1}{l}{sCIFAR10-20} & \multicolumn{1}{l}{sCIFAR10-40} & \multicolumn{1}{l}{sCIFAR10-80} & \multicolumn{1}{l}{sCIFAR10-160} & \multicolumn{1}{l}{sCIFAR10-320} & \multicolumn{1}{l}{sCIFAR10-640} & \multicolumn{1}{l}{sCIFAR10-1280} \\
\midrule

CNN-lc & \ding{55} & 0.0 & 0.271 & 0.324   &  0.361  &   0.418  &   0.453  &  0.501 & 0.541 & 0.594\\
CNN-lc & \ding{55} & 0.4 & 0.297  &  0.338  &  0.382  &  0.435  &  0.471  &  0.528 & 0.572  &  0.617\\
CNN-lc & \ding{55} & 0.7 & 0.297  &  0.349   &  0.400  &   0.449  &  0.494  &  0.539 & 0.581 & 0.623 \\
CNN-mc & \ding{55} & 0.0 & 0.285  &  0.343  &  0.388  &  0.430  &  0.486  &  0.537 & 0.588 & 0.633 \\
CNN-mc & \ding{55} & 0.4 & 0.297  &   0.349  &  0.399 & 0.454  &  0.509 & 0.558 & 0.610 & 0.662 \\
CNN-mc & \ding{55} & 0.7 & 0.315  &  0.362  &  0.413  &  0.471  &  0.519 & 0.577 & 0.628  &  0.676 \\
CNN-hc & \ding{55} & 0.0 & 0.301  &  0.342  &  0.391  &  0.447  &  0.506  &  0.561 & 0.612 & 0.659 \\
CNN-hc & \ding{55} & 0.4 & 0.317  &  0.361  &  0.408   & 0.465  &  0.520 & 0.580 & 0.635  &  0.688 \\
CNN-hc & \ding{55} & 0.7 & 0.319  &  0.370  &   0.425  &  0.481  &  0.539  &  0.595 & 0.648 & 0.698 \\
CNN-hc & \ding{51} & 0.7 & \textbf{0.325}  &  \textbf{0.382}  &   \textbf{0.455}   &  \textbf{0.526}  &  \textbf{0.603}  &  0.677 & 0.730 & 0.772 \\
ResNet-20 & \ding{55} & -- & 0.233  &  0.290  &  0.319 &  0.385  &  0.447  &   0.513 & 0.623 & 0.715\\
ResNet-20 & \ding{51} & -- & 0.266  &   0.339 & 0.385  &  0.469 & \textbf{0.603}  &   \textbf{0.710} & \textbf{0.794} & \textbf{0.848}\\

\midrule
& \multicolumn{9}{c}{\textbf{Dataset}}\\
\midrule
\textbf{Model}  & \multicolumn{1}{l}{Augmentation}  & \multicolumn{1}{c}{Dropout}  & \multicolumn{1}{l}{sFMNIST-10}  & \multicolumn{1}{l}{sFMNIST-20}  & \multicolumn{1}{l}{sFMNIST-40}  & \multicolumn{1}{l}{sFMNIST-80}  & \multicolumn{1}{l}{sFMNIST-160}  & \multicolumn{1}{l}{sFMNIST-320}  & \multicolumn{1}{l}{sFMNIST-640}  & \multicolumn{1}{l}{sFMNIST-1280}  \\
\midrule

CNN-lc & \ding{55} & 0.0 & 0.711 & 0.740  &  0.778 & 0.810 & 0.836 & 0.854 & 0.867 & 0.882 \\
CNN-lc & \ding{55} & 0.4 & 0.712  &  0.761   &  0.798 & 0.819  &  0.842  &  0.857 & 0.871 & 0.887  \\
CNN-lc & \ding{55} & 0.7 & 0.713  &  0.756  &  0.787  &  0.816  &  0.833  &  0.853 & 0.870  &  0.879 \\
CNN-mc & \ding{55} & 0.0 & 0.725  &  0.755 & 0.790   &  0.820  &  0.844 & 0.863  & 0.880  &  0.891 \\
CNN-mc & \ding{55} & 0.4 & 0.724 & 0.761  &  0.796  &  0.829 & 0.847  &   0.866 & 0.881 & 0.896 \\
CNN-mc & \ding{55} & 0.7 & 0.725  &  0.769  &  0.799   & 0.829 & 0.849  &  0.868 & 0.882 & 0.894 \\
CNN-hc & \ding{55} & 0.0 & 0.719  &  0.759  &  0.801 & 0.823  &  0.851  &  0.868 & 0.886  &   0.895 \\
CNN-hc & \ding{55} & 0.4 & 0.722  &  0.763  &  0.802  &  0.830   & 0.850 & 0.869 & 0.885  &  0.898\\

CNN-hc & \ding{55} & 0.7 & 0.733  &  0.774 &  0.805  &  0.832 & 0.853  &  0.871  & 0.888 & 0.899 \\
CNN-hc & \ding{51} & 0.7 & \textbf{0.739}  &  \textbf{0.777}  &  \textbf{0.812}  &  \textbf{0.844} & \textbf{0.865} & 0.880 & 0.896  &  0.908\\
ResNet-20 & \ding{55} & -- & 0.623 & 0.714  &  0.770 & 0.804  &  0.841 & 0.869 & 0.892 & 0.905 \\
ResNet-20 & \ding{51} & -- & 0.709 & 0.757  &  0.795  &  0.832  &  0.862  &  \textbf{0.890} & \textbf{0.905} & \textbf{0.919}\\
\midrule
& \multicolumn{9}{c}{\textbf{Dataset}}\\
\midrule
\textbf{Model}  & \multicolumn{1}{l}{Augmentation}  & \multicolumn{1}{c}{Dropout}  & \multicolumn{1}{l}{sSVHN-10}    & \multicolumn{1}{l}{sSVHN-20}    & \multicolumn{1}{l}{sSVHN-40}    & \multicolumn{1}{l}{sSVHN-80}    & \multicolumn{1}{l}{sSVHN-160}    & \multicolumn{1}{l}{sSVHN-320}    & \multicolumn{1}{l}{sSVHN-640}    & \multicolumn{1}{l}{sSVHN-1280}    \\
\midrule

CNN-lc & \ding{55} & 0.0 & 0.253 & 0.375 & 0.505 & 0.641 & 0.730 & 0.777 & 0.809 & 0.836 \\
CNN-lc & \ding{55} & 0.4 & 0.284 & 0.449 & 0.590 & 0.702  & 0.770 & 0.803 & 0.833 & 0.858 \\
CNN-lc & \ding{55} & 0.7 & 0.288 & 0.467 & 0.606 & 0.721 & 0.777 & 0.813  & 0.8323 & 0.862 \\
CNN-mc & \ding{55} & 0.0 & 0.268 & 0.399 & 0.534 & 0.686 & 0.755 & 0.796 & 0.832 & 0.855 \\
CNN-mc & \ding{55} & 0.4 & 0.296 & 0.433 & 0.643 & 0.721 & 0.783 & 0.822 & 0.848 & 0.873 \\
CNN-mc & \ding{55} & 0.7 & 0.277 & 0.458 & 0.641 & 0.746 & 0.797 & 0.832 & 0.862  & 0.882 \\
CNN-hc & \ding{55} & 0.0 & 0.249 &  0.375 & 0.555 & 0.677 & 0.745 & 0.801 & 0.840  & 0.861  \\
CNN-hc & \ding{55} & 0.4 & 0.275 & 0.456 & 0.631 & 0.736 & 0.793 & 0.827 & 0.857 & 0.881 \\

CNN-hc & \ding{55} & 0.7 & 0.288 & 0.448 & 0.647 & 0.744 & 0.795 & 0.840 & 0.863 & 0.886 \\
CNN-hc & \ding{51} & 0.7 & \textbf{0.344} & \textbf{0.582} &  \textbf{0.766} & \textbf{0.832} & 0.874  & 0.895 & 0.910   & 0.924 \\
ResNet-20 & \ding{55} & -- & 0.203 & 0.400 & 0.547 & 0.741 & 0.835 & 0.867 & 0.895 & 0.922 \\
ResNet-20 & \ding{51} & -- & 0.314 & 0.574  & 0.746  & 0.830  & \textbf{0.879} & \textbf{0.905} & \textbf{0.923}  & \textbf{0.938}  \\
\bottomrule                   
\end{tabular}

%% file: conclusion.tex
\section{Conclusions}
\label{sec:conclusions}

In this work, we have tackled the problem of image classification with few samples per class.
Although the results are still very far from the successes of high-capacity models with tons of data, we encourage the community to study and improve the capabilities of deep networks with tiny datasets.
We have run several experiments and trained different flavors of deep architectures on three popular computer vision benchmarks.\\
In the experiments regarding the influence of model complexity on performance, we have shown that relatively simple networks can play a major role in being less prone to overfitting and generalizing better when facing small datasets.
Therefore, architectures that will be proposed in the future, should not only be compared with state-of-the-art models (e.g. ResNet-34), but also with simpler networks.
On the other hand, when data augmentation is used, deeper networks rapidly gain higher performance even with basic augmentation policies thanks to the artificial addition of training images.\\
Furthermore, we run a wide analysis to test the benefits of using popular regularization techniques such as data augmentation and dropout in small data settings.
We figured out that the first one can induce large improvements in recognition capabilities depending on the dataset type and size. 
For what concerns dropout, we have shown that it generally improves results and should be used as well.
For future works, we would like to perform the proposed analysis on other datasets and implement more sophisticated data augmentation techniques.

\section*{Acknowledgement}
Work  partially  funded  by  the  EUHorizon   2020   research   and   innovation   program   under AI4EU  project  grant  N.  825619.